\documentclass{article} 
\usepackage{iclr2027_conference,times}
\iclrfinalcopy

\usepackage{amsmath,amssymb,bm}
\usepackage{graphicx}
\usepackage{booktabs}
\usepackage{xspace}
\usepackage{hyperref}
\usepackage{url}
\usepackage{cleveref}
\usepackage{colortbl}

\usepackage{tikz}
\usepackage{tcolorbox}

\newtcolorbox{researchquestion}{colback=cyan!3!white,colframe=black!65,boxrule=0.7pt,arc=6pt,boxsep=0pt,left=5pt,right=5pt,top=4pt,bottom=4pt,before skip=6pt,after skip=6pt,fontupper=\itshape}
\newcommand{\rqicon}{\tikz[baseline=-0.5ex,x=1pt,y=1pt]{\draw[line width=0.55pt,draw=black!65,fill=cyan!15] (0,0) circle[radius=3];\draw[line width=1.5pt,draw=black!65,line cap=round] (2.2,-2.2)--(5,-5);\draw[line width=0.65pt,draw=orange!45,line cap=round] (2.6,-2.6)--(4.7,-4.7);}}

\title{FutureWorld: A Plug-and-Play Causal Next-State Predictor for Interactive Video World Exploration}

\author{
  \vspace{-25pt}\\
  Guangyuan Li$^{1,2}$,\quad 
  Kelu Yao$^{3*}$, \quad
  Shuolin Xu$^{2,4}$,\quad 
  Xinlei Yu$^2$,\quad 
  Wei Xing$^1$,\quad
  Lei Zhao$^{1*}$,\quad
  \\
  Peng-Tao Jiang$^5$,\quad
  Xiaogang Xu$^1$,\quad
  Xiaobin Hu$^{2}$\thanks{Corresponding authors.}\vspace{3pt}
  \\
  $^1$ Zhejiang University 
  \quad
  $^2$ National University of Singapore
  \quad
  $^3$ Zhejiang Lab
  \\ 
  $^4$ Bournemouth University
  \quad 
  $^5$vivo BlueImage
}

\begin{document}

\maketitle

\begin{figure*}[h]
  \centering
  \includegraphics[width=\textwidth]{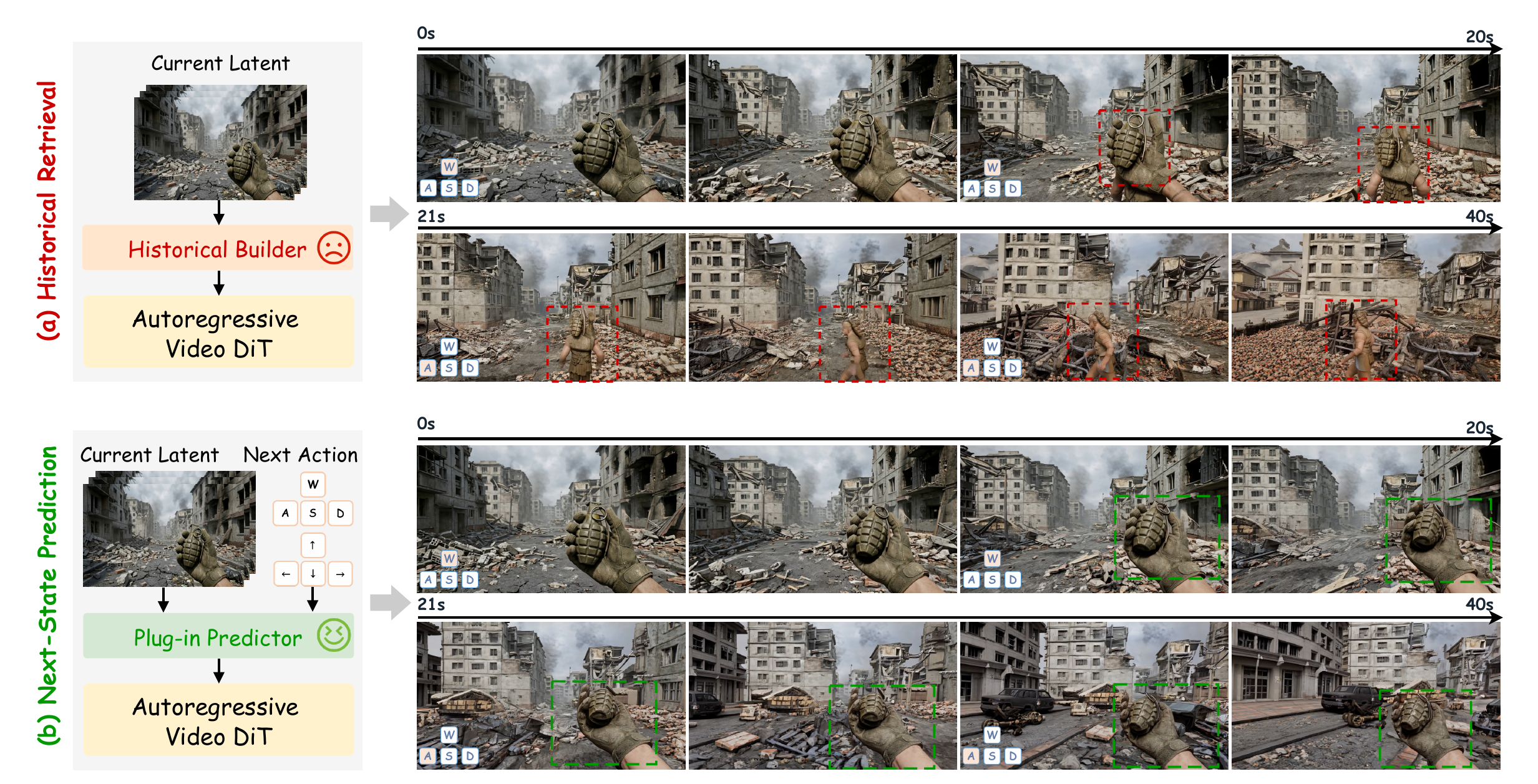}
  \caption{Improving long-horizon consistency through explicit next-state prediction. (a) Reusing generated history can carry existing errors into subsequent generations, allowing scene distortions and semantic drift to accumulate across interactions (red dashed boxes). (b) Our plug-in predictor estimates the next latent state from the current generated latent and the next action. This future-state prior guides the autoregressive video DiT through feature modulation, helping preserve scene structure and semantic consistency as interactions proceed (green dashed boxes).}
  \label{fig:teaser}
\end{figure*}

\begin{abstract}
Interactive video world models aim to follow user actions while maintaining scene consistency over extended interactions. History-based methods reuse generated latents to preserve scene content, but three limitations of this reuse contribute to \textbf{\emph{error recirculation}}. \textbf{1)} The latent generated at each interaction is retained as history, potentially containing blurred details, geometric distortions, or semantic drift. \textbf{2)} Although retrieved historical content matches the current scene, it can still contain errors that are carried into the next chunk. \textbf{3)} New outputs then become history for later interactions, potentially reinforcing the same errors through repeated reuse. To limit this cycle, we aim to predict the future scene state from the current generated latent and the next action before generating the next interaction's video chunk. This future-state prior provides guidance for the next chunk, with the aim of reducing the errors carried forward from history. To this end, we propose a plug-and-play Causal Next-Chunk Predictor (C-NCP) with cross-model adaptability and prediction-guided modulation. We integrate C-NCP into four base generators and evaluate the augmented models across three benchmarks for interactive long-video generation and action-conditioned world modeling. Extensive experiments demonstrate reduced error accumulation and improved long-horizon consistency across diverse generation settings.
The project page is at \url{https://GuangYuanKK.github.io/futureworld}.
\end{abstract}

\section{Introduction}
\label{sec:introduction}
Interactive video world models~\citep{li2026magicworld,sun2025worldplay,robbyant2026lingbotworld,wu2026infiniteworld,wang2026matrixgame3,zhu2026sanawm,zhao2026minwm,dreamx2026world,alayaworld2026alayaworld,chen2026reworld,zhao2026quantwm,ji2025sonic} generate visual environments that evolve under user action control. Sustained interaction requires each new chunk to follow the next action while preserving the scene's structure and semantics. As interactions continue, local generation defects can affect subsequent chunks, making error propagation a central obstacle to long-horizon consistency.

To maintain scene consistency, history-based approaches~\citep{sun2025worldplay,wang2026matrixgame3,li2025hunyuan,zhu2025memorize,yu2025context,zhao2026relax,yu2025videossm,kim2026memrope,yin2026framemorrow} repeatedly retrieve and reuse generated content. However, this reuse can lead to \emph{error recirculation}. Defects from earlier generations enter new outputs, which are then added to history and reused in later interactions, as illustrated in \cref{fig:teaser}. Three limitations of history reuse contribute to this cycle. \textbf{First}, the latent generated at each interaction is retained as history, potentially containing visual defects such as blurred details, geometric distortions, or semantic drift. \textbf{Second}, although a retrieved history view matches the current scene, it can still contain errors. Reusing such a view can therefore propagate these errors to the next chunk. \textbf{Third}, the new output is then added to history and reused in later interactions, potentially reinforcing the same errors. To limit this cycle, the generator needs guidance for the next chunk that does not rely solely on reusing potentially flawed history.

Based on this analysis, our motivation is to predict the future state from the current generated latent and the next action, providing an explicit future-state prior before generating the video chunk for the next interaction. Supervision from real video transitions teaches the predictor how scenes change under action control. It uses these learned transitions to re-estimate the upcoming state, rather than only selecting content already present in history. The resulting future-state prior guides the next generation, with the aim of reducing errors carried forward from earlier outputs. To turn this prior into effective guidance, we need to meet the following two requirements. \textbf{(1) Adaptability.} To make future-state prediction usable across different interactive video generators, the predictor must adapt to their latent representations. Different models use VAE latents with different dimensions and distributions, so the predictor needs a way to accommodate these differences. \textbf{(2) Modulation.} The future-state prior must also help guide the generation of the next chunk. Direct concatenation places the predicted latent alongside the generator's main latent, but the generator must still learn how to use this extra input when producing the next chunk. Modulation instead uses the future-state prior to adjust the main latent directly, while keeping its tensor shape unchanged. These considerations lead to our central research question.

\begin{researchquestion}
\rqicon\enspace How can we predict a future state from the current generated latent and the next action, and use it to guide subsequent generation and limit error propagation across interactions?
\end{researchquestion}

To address this question, we propose the plug-and-play Causal Next-Chunk Predictor (C-NCP), which estimates the complete next latent chunk from the current interaction's generated latent and the next interaction's action control. We use Bayesian decision theory ~\citep{sarkka2013bayesian} to formulate this prediction task as learning a future-state estimate that minimizes expected prediction loss given the current latent and next action. To enable adaptation, VAE-specific input adapters map each model's latents into a shared representation for the causal spatiotemporal predictor, while output heads map its predictions back to the corresponding model's latent space. To enable modulation, we use the future-state prior to adjust the generator's main latent through residual feature modulation based on adaptive instance normalization~\citep{huang2017adain}. After each interaction, the newly generated latent becomes the input to the next prediction, so the future-state prior is updated as interaction continues.
Our main contributions are summarized as follows.
\begin{itemize}
  \item We introduce a future-state-guided correction paradigm for error recirculation in interactive video generation. Rather than only selecting content from generated history, this approach learns to re-estimate the state expected under the next action. The resulting prior guides each new chunk before it becomes part of history.
  \item We develop the plug-and-play Causal Next-Chunk Predictor with cross-model adaptability and prediction-guided modulation. VAE-specific adapters let the predictor work with different latent representations, while residual modulation uses its future-state prior to adjust the generator's main latent.
  \item We integrate our C-NCP into four base generators and evaluate the augmented models across three benchmarks for interactive long-video generation and action-conditioned world modeling. Extensive experiments demonstrate reduced error accumulation and improved long-horizon consistency across diverse generation settings.
\end{itemize}

\section{Related Work}
\label{sec:related}

\noindent \textbf{Interactive Video World Models.}
Interactive video world models~\citep{li2026magicworld,sun2025worldplay,robbyant2026lingbotworld,wu2026infiniteworld,wang2026matrixgame3,zhu2026sanawm,zhao2026minwm,dreamx2026world,alayaworld2026alayaworld,chen2026reworld} extend visual generation to environments that evolve under action control. Early token-based systems such as iVideoGPT~\citep{wu2024ivideogpt} model observations, actions, and rewards within an autoregressive sequence. Matrix-Game 2.0~\citep{he2025matrixgame2} and LingBot-World~\citep{robbyant2026lingbotworld} develop real-time streaming generation across diverse environments, while SANA-WM~\citep{zhu2026sanawm} combines hybrid linear attention and long-video refinement for efficient minute-scale generation. Recent work also addresses the mismatch between clean training contexts and generated histories. Matrix-Game 3.0~\citep{wang2026matrixgame3} reuses prediction errors to expose the model to imperfect conditioning, and DreamX-World~\citep{dreamx2026world} combines residual recycling with training on self-generated long-horizon contexts. These approaches improve the generator's ability to continue from its own outputs rather than assuming error-free history. However, remaining defects in generated history can still affect subsequent chunks. Our plug-and-play Causal Next-Chunk Predictor provides an explicit future-state prior before the next interaction's video chunk is generated. It learns from real video transitions, and VAE-specific adapters accommodate different latent representations so that the predictor can be integrated into different base generators.

\noindent \textbf{History Modeling for Video Generation.}
History modeling preserves scene context through retrieval and compact memory. Spatial-memory world models~\citep{wu2025spatialmemory} reconstruct static 3D scenes for rendering along target trajectories, while WorldPlay~\citep{sun2025worldplay} reconstructs relevant historical context. WorldKV~\citep{yi2026worldkv} retrieves viewpoint-matched chunks and compresses redundant tokens; Matrix-Game 3.5~\citep{qian2026matrixgame35} uses geometry-aware patch memory; and ReWorld~\citep{chen2026reworld} combines a bounded KV cache with a pose-indexed landmark bank. Compression and adaptive selection further reduce context costs. FramePack~\citep{zhang2025framepack} uses frame-dependent compression and anti-drifting sampling, while TinyHistory~\citep{zhang2025tinyhistory} learns compact history embeddings through frame-query pretraining and content-consistency learning. Infinite-World~\citep{wu2026infiniteworld} uses pose-free hierarchical memory, and CaR~\citep{peng2026car} combines context compression with attention-based implicit retrieval. OmniMem~\citep{zhao2026omnimem} sparsely retrieves KV blocks from historical caches. Future Forcing~\citep{luo2026futureforcing} uses future-query proxies to guide KV retention and merging. Retrieval and compression help preserve scene content, but the generated latents they reuse can still contain visual defects. When these defects enter new outputs and those outputs become history, repeated reuse can cause error recirculation. To address this problem, our predictor learns how scenes change under action control from real video transitions and predicts the next latent state from the current generated latent and the next action. The resulting future-state prior guides the generator's main latent through modulation, aiming to reduce the errors passed into subsequent chunks.

\section{Method}
\label{sec:method}

\subsection{Overview}
\label{sec:overview}
\begin{figure*}[t]
  \centering
  \includegraphics[width=\textwidth]{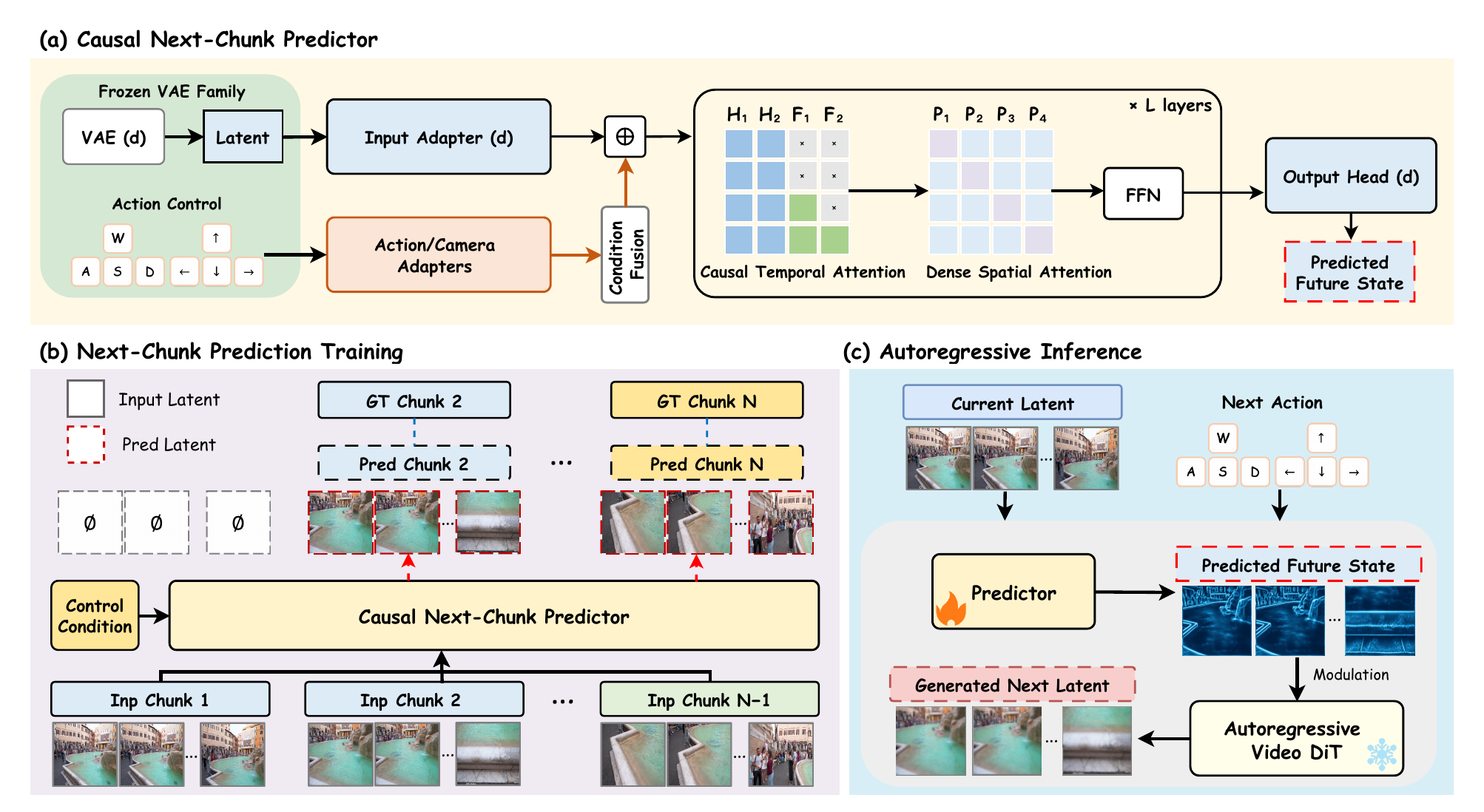}
  \vspace{-8mm}
  \caption{Overview of FutureWorld. Panel a presents the Causal Next-Chunk Predictor, which combines VAE-specific adapters, action conditioning, and a causal spatiotemporal backbone to predict the complete next latent chunk. Panel b illustrates next-chunk prediction training, where the predictor learns to forecast subsequent latent chunks from input chunks and control conditions under supervision from corresponding ground-truth chunks. Panel c shows autoregressive inference, where the predicted future state guides a frozen autoregressive video DiT through modulation to generate the next latent chunk.}
  \label{fig:method}
\end{figure*}

Reusing imperfect generated history can propagate existing defects into subsequent chunks, leading to error recirculation. We therefore learn a future-state prior from real video transitions to guide the next interaction's generation. As shown in \cref{fig:method}, the Causal Next-Chunk Predictor combines VAE-specific adapters for cross-model adaptability with a shared spatiotemporal trunk, learns through two-stage Next-Chunk Prediction Training, and uses prediction-guided Modulation to guide a frozen Autoregressive Video DiT during Autoregressive Inference. Prediction and generation follow
\begin{equation}
  \widehat{\mathbf{z}}_{t+1}^{(d)}=P_\phi^{(d)}(\widetilde{\mathbf{z}}_t^{(d)},\mathbf{u}_{t+1}),
  \qquad
  \widetilde{\mathbf{z}}_{t+1}^{(d)}=G_\theta^{(d)}\!\left(\mathcal{H}_t,\mathbf{u}_{t+1};\widehat{\mathbf{z}}_{t+1}^{(d)}\right),
  \label{eq:predictor-overview}
\end{equation}
where $t$ indexes interaction chunks and $d$ identifies the VAE domain. The latents $\widetilde{\mathbf{z}}_t^{(d)}$, $\widehat{\mathbf{z}}_{t+1}^{(d)}$, and $\widetilde{\mathbf{z}}_{t+1}^{(d)}$ are the current generated state, predicted next state, and completed next output, respectively. Each lies in $\mathbb{R}^{C_d\times T_d\times H_d\times W_d}$, with $C_d$ channels, $T_d$ temporal slices, height $H_d$, and width $W_d$. Here $\mathbf{u}_{t+1}$ is the next action, $\mathcal{H}_t$ is the generated history, and $P_\phi^{(d)}$ and $G_\theta^{(d)}$ are the predictor and generator with trainable parameters $\phi$ and frozen parameters $\theta$. The semicolon denotes prediction-guided modulation. Completed outputs update history and feed subsequent predictions; real-video latents are denoted by $\mathbf{z}_t^{(d)}$. 

\subsection{Causal Next-Chunk Predictor}
\label{sec:predictor}
Errors in the current generated latent may propagate into subsequent generation. We therefore supervise the predictor with real future states to learn action-induced scene changes. To match sequential forecasting, we condition each future slice only on the current input, the next action, and preceding slices. The probability chain rule gives
\begin{equation}
  p_*(\mathbf{Y}\mid C)
  =\prod_{\tau=1}^{T_d}
    p_*(\mathbf{Y}_\tau\mid\mathbf{Y}_{<\tau},C),
  \label{eq:predictive-chain}
\end{equation}
where $p_*$ is the data distribution, $C=(\mathbf{x}_t,\mathbf{u}_{t+1})$ is the conditioning information, and $\mathbf{x}_t$ is the predictor input, equal to $\widetilde{\mathbf{z}}_t$ at inference. The random tensor $\mathbf{Y}$ is the real next chunk, $\mathbf{Y}_\tau$ is its slice at time $\tau$, and $\mathbf{Y}_{<\tau}$ denotes preceding slices. Rather than estimating a density, we learn a point estimate, motivated by Bayesian decision theory~\citep{sarkka2013bayesian}
\begin{equation}
  a_\ell^*(C)\in\underset{\mathbf{a}}{\operatorname{argmin}}\,
  \mathbb{E}_{p_*}\!\left[\ell(\mathbf{a},\mathbf{Y})\mid C\right],
  \label{eq:bayes-decision}
\end{equation}
where $\mathbf{a}$ is a candidate future latent, $\ell$ is its prediction loss, and $a_\ell^*(C)$ is an optimal estimate. The operator $\operatorname{argmin}$ selects minimizers and $\mathbb{E}_{p_*}[\cdot\mid C]$ denotes conditional expectation; expectations use $p_*$ throughout. With squared error and finite second moments, the optimum is
\begin{equation}
  m(C)=\mathbb{E}[\mathbf{Y}\mid C]
  \in\underset{\mathbf{a}}{\operatorname{argmin}}\,
    \mathbb{E}\!\left[\|\mathbf{Y}-\mathbf{a}\|_F^2\mid C\right],
  \label{eq:conditional-projection}
\end{equation}
where $m(C)$ is the conditional mean and $\|\cdot\|_F$ is the Frobenius norm. Its orthogonal-projection interpretation~\citep{dembo2021probability} yields
\begin{equation}
  \mathbb{E}\|\mathbf{Y}-f(C)\|_F^2
  =\underbrace{\mathbb{E}\|\mathbf{Y}-m(C)\|_F^2}_{\text{conditional uncertainty}}
   +\underbrace{\mathbb{E}\|m(C)-f(C)\|_F^2}_{\text{excess prediction risk}},
  \label{eq:risk-decomposition}
\end{equation}
where $f(C)$ is any square-integrable predictor and expectations average over the joint distribution of $C$ and $\mathbf{Y}$. This separates conditional uncertainty from reducible prediction error, motivating a learned future-state prior. The analysis concerns squared error, not the exact optimum of our normalized mixed loss or a guarantee of rollout error contraction; Appendix~\ref{app:decision-details} provides the derivation.

To enable cross-model adaptability, we account for differences in VAE latent channel counts and distributions. Input Adapter (d) learns a $1\times1\times1$ channel mapping to a shared hidden space, preserving the temporal/spatial grid while accommodating these differences. Action/Camera Adapters and Condition Fusion produce
\begin{equation}
  \mathbf{H}^{(0)}
  =\left[I_d(\mathbf{x}_t)\,;\,\mathbf{Q}+\Pi A_d(\mathbf{u}_{t+1})\right]+\mathbf{E},
  \label{eq:condition-fusion}
\end{equation}
where $\mathbf{H}^{(0)}$ is the trunk input, $I_d$ is Input Adapter (d), $A_d$ encodes action control, and $\Pi$ aligns control features to the temporal/spatial query grid. The queries $\mathbf{Q}$ encode a learned start token followed by preceding ground-truth slices during training or predicted slices during forecasting. Here $\mathbf{E}$ contains type/domain embeddings and $[\,;\,]$ concatenates temporal positions. Causal Temporal Attention uses
\begin{equation}
  M_{ij}=0\ \text{if }j\in\mathcal{I}_{H}\ \text{or }(i=F_k,\ j=F_r,\ r\leq k),\qquad M_{ij}=-\infty\ \text{otherwise}.
  \label{eq:causal-mask}
\end{equation}
where $M_{ij}$ is the mask entry for query $i$ and key $j$, $\mathcal{I}_{H}$ contains current-input positions, and $F_k,F_r$ are future-query positions with $1\leq k,r\leq T_d$. Zero permits attention and $-\infty$ blocks it. Shifted queries prevent access to their own or later targets. The trunk combines Causal Temporal Attention, Dense Spatial Attention, and feed-forward networks; Output Head (d) returns
\begin{equation}
  \widehat{\mathbf{z}}_{t+1}^{(d)}
  =O_d\!\left(\left[T_\phi^{(L)}(\mathbf{H}^{(0)};\mathbf{M})\right]_{F}\right),
  \label{eq:direct-output}
\end{equation}
where $T_\phi^{(L)}$ is the shared trunk of $L$ blocks, $\mathbf{M}=(M_{ij})$, $[\cdot]_F$ selects future queries, and $O_d$ maps features to native VAE channels. The output is a complete latent, not an input residual. Block details are in Appendix~\ref{app:architecture-details}.

\subsection{Next-Chunk Prediction Training}
\label{sec:training}

Since C-NCP is trained from scratch, its early predictions are unreliable, and directly feeding them back across chunks can destabilize optimization. We therefore adopt a two-stage training schedule that first supports early learning with ground-truth-assisted inputs and then adapts the predictor to prediction-only inter-chunk feedback. Stage I adds aligned predicted and ground-truth chunks element-wise, retaining predicted content while providing ground-truth support. Stage II removes the direct ground-truth addition and continues optimizing the same predictor. Both stages retain within-chunk teacher forcing and start from a real initial chunk, with subsequent inter-chunk inputs constructed as follows
\begin{equation}
  \begin{aligned}
    \mathbf{x}_1^{(s)}&=\mathbf{z}_1,\qquad
    \widehat{\mathbf{z}}_{n+1}^{(s)}=P_\phi^{\mathrm{TF}}(\mathbf{x}_n^{(s)},\mathbf{u}_{n+1};
      \operatorname{shift}(\mathbf{z}_{n+1})),\\
    \mathbf{x}_n^{(s)}&=
      \begin{cases}
        \widehat{\mathbf{z}}_n^{(1)}+\mathbf{z}_n,&s=1,\\
        \widehat{\mathbf{z}}_n^{(2)},&s=2,
      \end{cases}
      \qquad n=2,\ldots,N-1,
  \end{aligned}
  \label{eq:add-feedback}
\end{equation}
Here, $N \geq 3$ is the number of training chunks, $n$ indexes chunks, and $s \in \{1,2\}$ denotes the training stage. The tensors $\mathbf{x}_n^{(s)}$, $\widehat{\mathbf{z}}_n^{(s)}$, and $\mathbf{z}_n$ denote the input, predicted chunk, and real target, respectively, while $\mathbf{u}_{t+1}$ specifies the next action. For $n = 1,\ldots,N-1$, the operator $P_\phi^{\mathrm{TF}}$ performs within-chunk teacher-forced prediction, and $\operatorname{shift}$ prepends the learned start token and removes the final target slice. Thus, Stage II passes predicted chunks to subsequent prediction steps, while those chunks are still computed using ground-truth prefixes within each chunk. Both stages supervise predictions against real future chunks.

State matching alone does not explicitly supervise temporal changes. Both training stages combine normalized state losses with adjacent-slice difference supervision
\begin{equation}
  (\Delta_T\mathbf{e}_n^{(s)})_\tau
  =(\widehat{\mathbf{z}}_{n,\tau}^{(s)}-\widehat{\mathbf{z}}_{n,\tau-1}^{(s)})
   -(\mathbf{z}_{n,\tau}-\mathbf{z}_{n,\tau-1}),
  \quad \tau=2,\ldots,T_d,
  \label{eq:ltemp}
\end{equation}
where $\mathbf{e}_n^{(s)}=\widehat{\mathbf{z}}_n^{(s)}-\mathbf{z}_n$ is the prediction error at stage $s\in\{1,2\}$, $\Delta_T$ takes temporal first differences, and the subscript $\tau$ selects a slice. This matches changes rather than suppressing motion. At each stage, we optimize
\begin{equation}
  \mathcal{L}^{(s)}(\phi)=\frac{1}{N-1}\sum_{n=2}^{N}
  \left[
    \lambda_1\left\langle\left|\frac{\mathbf{e}_n^{(s)}}{s_n}\right|\right\rangle
    +\lambda_2\left\langle\left(\frac{\mathbf{e}_n^{(s)}}{s_n}\right)^2\right\rangle
    +\lambda_{\mathrm{temp}}\left\langle\left|\frac{\Delta_T\mathbf{e}_n^{(s)}}{r_n}\right|\right\rangle
  \right],
  \label{eq:loss}
\end{equation}
where $\mathcal{L}^{(s)}(\phi)$ is the loss at stage $s$ and $\lambda_1,\lambda_2,\lambda_{\mathrm{temp}}\geq0$ weight absolute, squared, and temporal-difference errors. Brackets $\langle\cdot\rangle$ average element-wise quantities over all tensor entries; $s_n$ and $r_n$ are positive normalization factors computed from target-state and target-difference standard deviations, respectively. They depend only on real targets and are shared across stages; Appendix~\ref{app:normalized-training} gives their definitions. Training updates only the predictor and its adapters.

\subsection{Autoregressive Inference}
\label{sec:modulation}
At each interaction, C-NCP receives the completed generator output $\widetilde{\mathbf{z}}_t^{(d)}$ and the next action $\mathbf{u}_{t+1}$. It predicts the next latent chunk $\widehat{\mathbf{z}}_{t+1}^{(d)}$ autoregressively, starting from the learned start token and using preceding predicted slices for subsequent queries. No future ground-truth slices are used. The predicted chunk then provides a future-state prior for the frozen video generator.
To incorporate this prior without enlarging the generator's main latent through concatenation, we use shape-preserving channel-wise modulation inspired by FiLM~\citep{perez2018film} and AdaIN~\citep{huang2017adain}. Rather than directly matching reference statistics, we derive residual scale and shift adjustments from the predicted and current states.
\begin{equation}
  \gamma_c=\frac{\sigma_c(\widehat{\mathbf{z}}_{t+1}^{(d)})}{\sigma_c(\widetilde{\mathbf{z}}_t^{(d)})}-1,
  \qquad
  \beta_c=\mu_c(\widehat{\mathbf{z}}_{t+1}^{(d)})-\mu_c(\widetilde{\mathbf{z}}_t^{(d)}),
  \label{eq:scale-shift}
\end{equation}
where $c\in\{1,\ldots,C_d\}$ indexes channels, $\gamma_c$ is the relative scale adjustment, and $\beta_c$ is the mean shift. Channel means $\mu_c$ and population standard deviations $\sigma_c$ use temporal/spatial entries, with standard deviations floored at $\epsilon>0$. During synthesis, $\widetilde{\mathbf{z}}_{t+1}^{(d)}$ denotes the working main latent, updated by
\begin{equation}
  \widetilde{\mathbf{z}}_{t+1,c}^{(d)}
  \leftarrow\widetilde{\mathbf{z}}_{t+1,c}^{(d)}
  +\alpha\!\left(
    \gamma_c\frac{\widetilde{\mathbf{z}}_{t+1,c}^{(d)}-\mu_c(\widetilde{\mathbf{z}}_{t+1}^{(d)})}
      {\sigma_c(\widetilde{\mathbf{z}}_{t+1}^{(d)})}
    +\beta_c\right),
  \label{eq:modulation}
\end{equation}
Here, $\leftarrow$ denotes an in-place update, and $\alpha \geq 0$ controls the modulation strength. The channel-wise coefficients are broadcast over the temporal and spatial dimensions. After synthesis, the completed generator output $\widetilde{\mathbf{z}}_{t+1}^{(d)}$, rather than the predicted prior $\widehat{\mathbf{z}}_{t+1}^{(d)}$, updates the history and becomes the input to the next prediction. Both networks remain frozen during inference.

\section{Experiments}
\label{sec:experiments}

\subsection{Datasets and Metrics}
\label{sec:datasets}
We train the Causal Next-Chunk Predictor on Sekai~\citep{li2025sekai} and evaluate on Sekai-val, RealEstate10K-test~\citep{zhou2018stereomag}, and the official WBench~\citep{ying2026wbench} test set. All evaluation videos are 60 seconds long. We report six VBench~\citep{huang2024vbench} dimensions and their average on Sekai-val and RealEstate10K-test, and five WBench dimensions and their average on WBench. These metrics cover visual quality, temporal and scene consistency, and interactive behavior; metric names are defined in the table captions.

\subsection{Implementation Details}
\label{sec:implementation}
Each training sequence contains 1800 frames divided into 60-frame chunks. We train the predictor on eight A100 GPUs following the two-stage schedule in \cref{sec:training}, with 100k iterations per stage. The predictor and its adapters are optimized using AdamW with a learning rate of $10^{-4}$. The normalized loss uses $\lambda_1=1$, $\lambda_2=1$, and $\lambda_{\mathrm{temp}}=0.25$. The VAEs and base generators remain frozen. Architecture and additional optimization settings are provided in Appendices~\ref{app:architecture-details} and~\ref{app:normalized-training}.
We use four base generators: Matrix-Game 3.0~\citep{wang2026matrixgame3}, HY-World 1.5~\citep{hunyuanworld2025hyworld15}, LingBot-World-Fast~\citep{robbyant2026lingbotworld}, and SANA-WM~\citep{zhu2026sanawm}.


\begin{table*}[t]
\centering
\caption{Quantitative comparison on RealEstate10K-test using VBench. All videos are 60 seconds long. Higher is better for every metric. Bold rows denote base generators augmented with our predictor. T.F.: Temporal Flickering; M.S.: Motion Smoothness; S.C.: Subject Consistency; B.C.: Background Consistency; A.Q.: Aesthetic Quality; I.Q.: Imaging Quality; Avg.: Average.}
\label{tab:vbench_realestate10k}
\begingroup
\resizebox{\textwidth}{!}{%
\begin{tabular}{l|cccccc|c}
\hline
\rowcolor{gray!18}
Method & T.F. $\uparrow$ & M.S. $\uparrow$ & S.C. $\uparrow$ & B.C. $\uparrow$ & A.Q. $\uparrow$ & I.Q. $\uparrow$ & Avg. $\uparrow$ \\
\hline
YUME 1.5 & 0.9127 & 0.9151 & 0.6928 & 0.9036 & 0.4128 & 0.7132 & 0.7584 \\
FantasyWorld & 0.9149 & 0.9128 & 0.6975 & 0.9020 & 0.4149 & 0.7104 & 0.7588 \\
Infinite-World & 0.9112 & 0.9170 & 0.6902 & 0.9067 & 0.4116 & 0.7160 & 0.7588 \\
\hline
SANA-WM & 0.9246 & 0.9764 & 0.7994 & 0.9034 & 0.5050 & 0.7401 & 0.8081 \\
\textbf{SANA-WM + Ours} & \textbf{0.9366} & \textbf{0.9867} & \textbf{0.8437} & \textbf{0.9202} & \textbf{0.5192} & \textbf{0.7532} & \textbf{0.8266} \\
\hline
Matrix-game-3.0 & 0.9241 & 0.9748 & 0.7946 & 0.9042 & 0.5025 & 0.7740 & 0.8124 \\
\textbf{Matrix-game-3.0 + Ours} & \textbf{0.9361} & \textbf{0.9884} & \textbf{0.8387} & \textbf{0.9210} & \textbf{0.5167} & \textbf{0.7877} & \textbf{0.8314} \\
\hline
HY-World 1.5 & 0.9238 & 0.9762 & 0.7965 & 0.9058 & 0.5040 & 0.7455 & 0.8086 \\
\textbf{HY-World 1.5 + Ours} & \textbf{0.9358} & \textbf{0.9898} & \textbf{0.8407} & \textbf{0.9226} & \textbf{0.5182} & \textbf{0.7587} & \textbf{0.8276} \\
\hline
LingBot-World-Fast & 0.9259 & 0.9783 & 0.8070 & 0.9002 & 0.5084 & 0.7009 & 0.8034 \\
\textbf{LingBot-World-Fast + Ours} & \textbf{0.9380} & \textbf{0.9819} & \textbf{0.8518} & \textbf{0.9169} & \textbf{0.5227} & \textbf{0.7133} & \textbf{0.8208} \\
\hline
\end{tabular}%
}
\endgroup
\vspace{-4mm}
\end{table*}

\subsection{Quantitative Comparison}
\label{sec:sota}
We compare all four base generators with their predictor-augmented counterparts and include other interactive video world models as external references. Sekai-val and WBench results are provided in Appendices~\ref{app:sekai-results} and~\ref{app:wbench-results}.
On RealEstate10K-test as shown in \cref{tab:vbench_realestate10k}, every predictor-augmented generator improved across all reported metrics. Subject and background consistency increased alongside temporal and visual-quality scores, with Matrix-Game 3.0 + Ours achieving the highest average. The improvements across all four bases indicate that the benefit is not confined to a single generator. Since the predictor was trained on Sekai, these results also support cross-dataset transfer of the learned transition prior. The gains are consistent with explicit future-state prediction providing a useful reference for maintaining scene continuity under action-driven viewpoint changes.

\subsection{Qualitative Comparison}

\begin{figure*}[h]
  \centering
  \includegraphics[width=\textwidth]{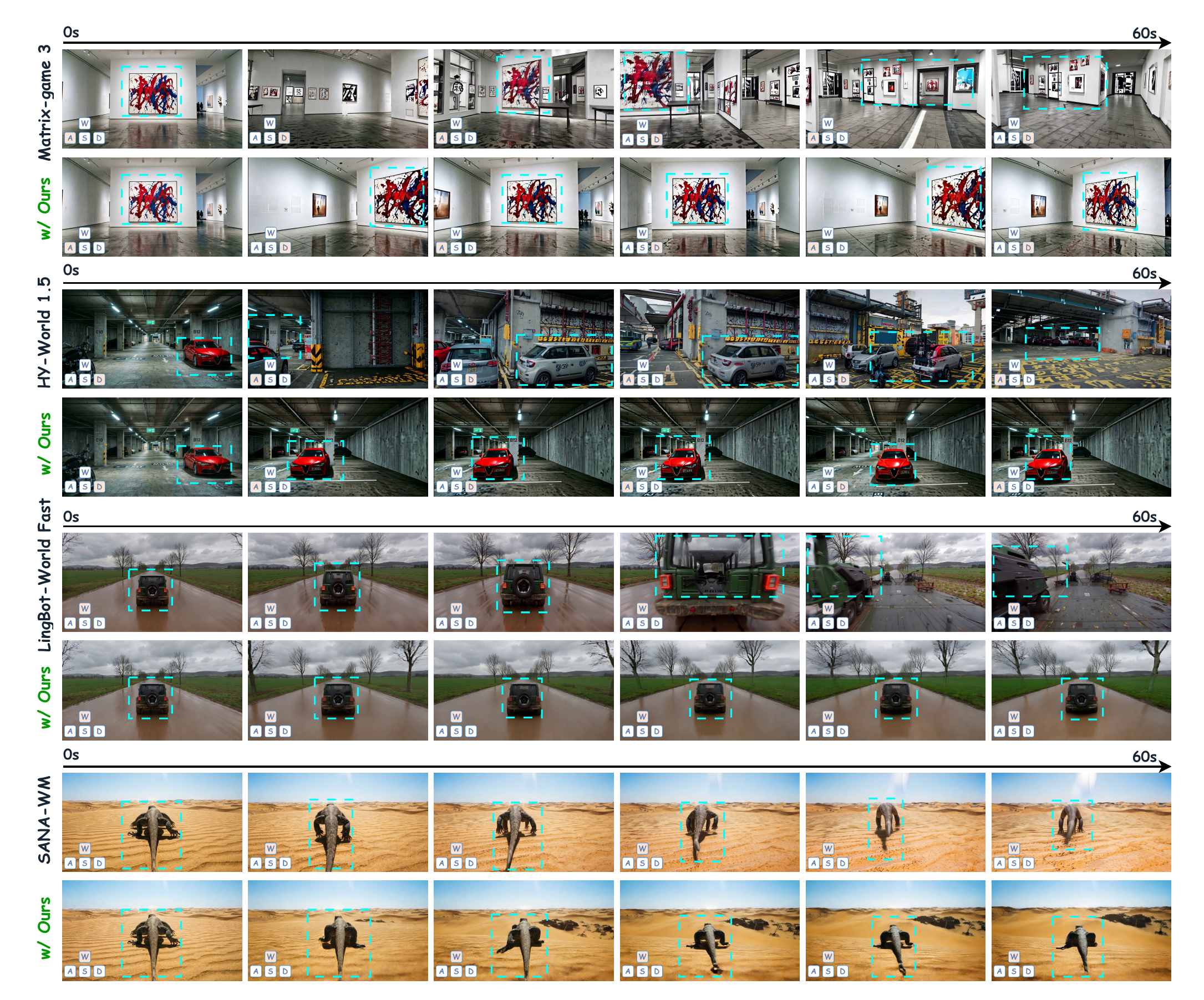}
  \vspace{-8mm}
  \caption{Qualitative comparisons over 60-second interactive rollouts with four base generators. Each pair shows the original generator above and its predictor-augmented counterpart below, with six frames ordered from left to right. Action controls appear in the lower-left corners, and cyan dashed boxes highlight subjects and scene regions for comparison. The examples show improved preservation of subject appearance and scene layout with our predictor.}
  \label{fig:qualitative}
  \vspace{-6mm}
\end{figure*}

\Cref{fig:qualitative} compares four base models with and without our predictor across the sampled 60-second rollouts. For Matrix-Game 3.0, our predictor better preserved the prominent painting and gallery layout, which changed in the baseline. For HY-World 1.5, it maintained the red car and garage structure, while the baseline drifted toward a different vehicle-filled setting. For LingBot-World-Fast, it retained vehicle appearance and road layout, whereas later baseline frames showed vehicle distortion and disrupted road content. For SANA-WM, it preserved clearer elephant head and ear contours than the baseline in later frames.
These examples illustrate improved subject and scene consistency during continued interaction, in recognizable object details and the surrounding scene layout. Each forecast uses the current generated latent and the next action, so the reference follows the ongoing interaction. The reduced drift is consistent with this supervised future-state reference guiding each new chunk through modulation, helping refine generation beyond the reuse of potentially imperfect history.
User-study results are provided in Appendix~\ref{app:user-study}.

\begin{figure*}[t]
  \centering
  \includegraphics[width=\textwidth]{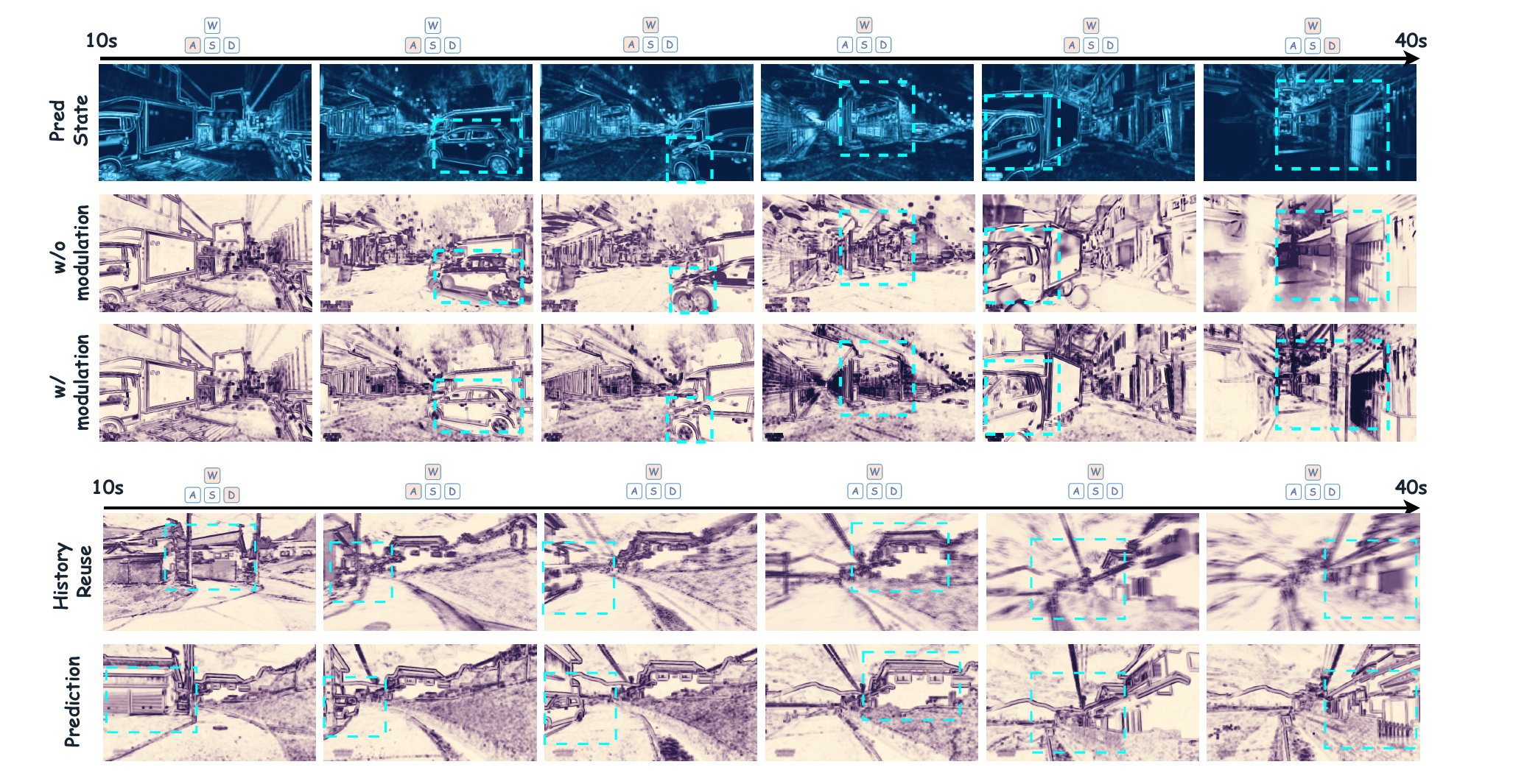}
  \vspace{-8mm}
  \caption{The upper panel compares the generator's main latent before and after modulation, with the predicted future state shown in blue. The lower panel compares main latents obtained with historical retrieval and predictor-guided modulation.}
  \label{fig:ablation-latents}
\end{figure*}

\begin{table*}[t]
\centering
\caption{Component ablation of the predictor with Matrix-game-3.0 as the base generator on RealEstate10K-test using VBench. All videos are 60 seconds long. Higher is better for every metric. The full method is highlighted in bold. Causal: causal prediction; Cond.: control conditioning; Temp.: temporal loss; Concat: direct concatenation; Mod.: modulation.}
\label{tab:predictor_ablation}
\begingroup
\resizebox{\textwidth}{!}{%
\begin{tabular}{l|ccccc|cccccc|c}
\hline
\rowcolor{gray!18}
Variant & Causal & Cond. & Temp. & Concat & Mod. & T.F. $\uparrow$ & M.S. $\uparrow$ & S.C. $\uparrow$ & B.C. $\uparrow$ & A.Q. $\uparrow$ & I.Q. $\uparrow$ & Avg. $\uparrow$ \\
\hline
Baseline & $\times$ & $\times$ & $\times$ & $\times$ & $\times$ & 0.9241 & 0.9748 & 0.7946 & 0.9042 & 0.5025 & 0.7740 & 0.8124 \\
Bidirectional Predictor & $\times$ & $\checkmark$ & $\checkmark$ & $\times$ & $\checkmark$ & 0.9307 & 0.9823 & 0.8189 & 0.9134 & 0.5103 & 0.7815 & 0.8229 \\
\emph{w/o} Condition & $\checkmark$ & $\times$ & $\checkmark$ & $\times$ & $\checkmark$ & 0.9295 & 0.9789 & 0.8144 & 0.9113 & 0.5103 & 0.7809 & 0.8209 \\
\emph{w/o} Temporal Loss & $\checkmark$ & $\checkmark$ & $\times$ & $\times$ & $\checkmark$ & 0.9283 & 0.9789 & 0.8312 & 0.9180 & 0.5139 & 0.7850 & 0.8259 \\
Direct Concat & $\checkmark$ & $\checkmark$ & $\checkmark$ & $\checkmark$ & $\times$ & 0.9332 & 0.9850 & 0.8300 & 0.9180 & 0.5110 & 0.7848 & 0.8270 \\
\hline
\textbf{Full Ours} & $\checkmark$ & $\checkmark$ & $\checkmark$ & $\times$ & $\checkmark$ & \textbf{0.9361} & \textbf{0.9884} & \textbf{0.8387} & \textbf{0.9210} & \textbf{0.5167} & \textbf{0.7877} & \textbf{0.8314} \\
\hline
\end{tabular}%
}
\endgroup
\vspace{-5mm}
\end{table*}

\subsection{Ablation Studies}
\label{sec:ablations}

\Cref{tab:predictor_ablation} compares the full model against five variants: the baseline without a predictor, bidirectional prediction, removal of action conditioning, removal of temporal loss, and direct concatenation instead of modulation. The full model achieved the highest score on every metric, while all predictor-equipped variants exceeded the baseline, supporting the value of an explicit future-state reference.

\noindent\textbf{Effectiveness of Causal Prediction and Action Conditioning.} Replacing bidirectional prediction with causal prediction improved temporal stability, subject consistency, and visual quality. The causal mask restricts future queries to available context and preceding slices, aligning prediction with the temporal direction of generation. Removing action conditioning yielded the lowest average and subject-consistency scores among predictor-equipped variants. Action control specifies which transition should follow the current latent, helping the predictor anticipate the intended scene change rather than infer it from appearance alone.

\noindent\textbf{Effectiveness of Temporal Supervision and Feature Fusion.} Adding temporal supervision improved flickering, motion smoothness, and scene consistency. Supervising differences between adjacent latent slices encourages coherent transitions rather than accurate individual states alone. Direct concatenation also improved over the baseline, but modulation outperformed it on every metric. Through channel-wise scale and shift, modulation lets the future-state prior refine the generator's existing main latent without replacing it. This comparison supports modulation as an effective way to incorporate the predicted reference during synthesis. Training-schedule ablations are reported in Appendix~\ref{app:training-strategy-ablation}.

\noindent\textbf{Effectiveness of Prediction-Guided Modulation.} The upper panel of \cref{fig:ablation-latents} shows recognizable vehicle contours and street structure in the predicted state, providing a visual reference for the subsequent modulation. Without modulation, the main latent exhibits fragmented vehicle boundaries and indistinct building details; modulation produces more continuous vehicle outlines, facades, and street boundaries. These observations are consistent with the scale-and-shift update in \cref{eq:modulation} using the future-state reference to refine the generator's existing scene structure during synthesis.

\noindent\textbf{History Reuse vs. Prediction.} In the lower panel of \cref{fig:ablation-latents}, main latents obtained with historical retrieval exhibit scene drift and blurring, particularly in later snapshots as interaction continues. Predictor-guided modulation preserves more stable scene structure and clearer building and road contours. A forecast learned from real transitions offers an additional action-conditioned reference when the reused history contains defects. These observations illustrate its benefit for refining generation on top of history reuse, where existing defects can otherwise persist in subsequent outputs.

\subsection{Inference-Time Analysis}
\label{sec:inference-time}

\begin{table}[htbp]
\begin{minipage}[t]{0.47\linewidth}
\vspace{\abovecaptionskip}
\Cref{tab:inference_time} quantifies the additional inference time introduced by the predictor. Across four interactions, the total generation time increased from 26.707\,s to 29.174\,s, while the average time per interaction increased from 6.677\,s to 7.294\,s. This corresponds to an additional 0.617\,s per interaction, or 9.24\% relative to the baseline. The predictor therefore provides explicit future-state guidance with a modest increase in generation time.
\end{minipage}\hfill
\begin{minipage}[t]{0.51\linewidth}
\vspace{0pt}
\centering
\caption{Inference-time comparison of Matrix-game-3.0 with and without the predictor when generating four 720P video chunks on an A800 GPU. Avg. denotes the time per interaction.}
\label{tab:inference_time}
\begin{tabular}{l|cc}
\hline
\rowcolor{gray!18}
Metric & Baseline & \textbf{\emph{w/} Predictor} \\
\hline
Total (s) & 26.707 & 29.174 \\
Avg. (s/interaction) & 6.677 & 7.294 \\
Extra (s/interaction) & -- & 0.617 \\
Overhead & -- & 9.24\% \\
\hline
\end{tabular}
\end{minipage}
\end{table}
\vspace{-4mm}

\section{Conclusion}
\label{sec:conclusion}
We presented the Causal Next-Chunk Predictor, a plug-and-play module that guides interactive video generation through explicit future-state prediction. Trained through direct supervision from real video transitions, it estimates the next latent chunk from the current interaction's generated latent and the next action control. Feature modulation incorporates this learned reference into the generator's main latent to refine each new chunk before it becomes part of the history. Experiments with four base generators on Sekai-val, RealEstate10K-test, and WBench show consistent improvements in visual quality and long-horizon consistency over the corresponding baselines. These findings support future-state prediction as an effective guide for more coherent generation in existing world models during sustained interaction.

\section*{AI Use Statement}
In this work, we used generative AI tools to improve the clarity, grammar, and presentation of the manuscript. All AI-assisted content was carefully reviewed and verified by the authors. All scientific ideas, methodological designs, experimental results, and conclusions were developed and validated by the authors. We take full responsibility for the final content of this work.

\bibliography{iclr2027_conference}
\bibliographystyle{iclr2027_conference}

\clearpage
\appendix
\raggedbottom
\setcounter{table}{0}
\renewcommand{\thetable}{A\arabic{table}}
\renewcommand{\theHtable}{appendix.\arabic{table}}

\section*{Summary of the Appendix}
This supplementary material provides additional quantitative results, analyses, and method details for \emph{FutureWorld: A Plug-and-Play Causal Next-State Predictor for Interactive Video World Exploration}. The appendix is organized as follows.
\begin{itemize}
  \item \S\ref{app:method-details} gives the decision-theoretic derivation, predictor architecture, and normalized training objective.
  \item \S\ref{app:sekai-results} reports VBench results on Sekai-val and analyzes the gains across four base generators.
  \item \S\ref{app:wbench-results} reports results on the official WBench test set and analyzes interactive generation quality and consistency.
  \item \S\ref{app:training-strategy-ablation} compares one-stage and two-stage training strategies for the predictor.
  \item \S\ref{app:additional-qualitative} presents additional qualitative comparisons in desert and coastal scenes.
  \item \S\ref{app:user-study} presents a user study evaluating perceptual preferences across four base generators.
\end{itemize}

\section{Additional Method Details}
\label{app:method-details}

\subsection{Conditional-Mean Interpretation}
\label{app:decision-details}

The squared-error interpretation in \cref{eq:conditional-projection,eq:risk-decomposition} separates uncertainty in the future state from error in its estimate. For finite-second-moment $\mathbf{Y}$ and a square-integrable predictor $f(C)$, write $\mathbf{Y}-f(C)=(\mathbf{Y}-m(C))+(m(C)-f(C))$. The cross term vanishes by conditional expectation
\begin{equation*}
  \begin{aligned}
    &\mathbb{E}\langle\mathbf{Y}-m(C),m(C)-f(C)\rangle_F\\
    &\quad=\mathbb{E}\langle\mathbb{E}[\mathbf{Y}-m(C)\mid C],m(C)-f(C)\rangle_F=0,
  \end{aligned}
\end{equation*}
where $\langle\cdot,\cdot\rangle_F$ is the Frobenius inner product over latent entries and $m(C)=\mathbb{E}[\mathbf{Y}\mid C]$. Expanding the squared norm then gives \cref{eq:risk-decomposition}. The first term measures uncertainty remaining under the available context; the second measures deviation from its conditional mean. This motivates learning a future-state reference from real transitions, but does not establish error contraction on generated inputs. The normalized mixed objective in \cref{eq:loss}, which also includes absolute-error and temporal terms, need not share the conditional-mean optimum. Finite-capacity training and differences between training and generated contexts further limit this idealized interpretation.

\subsection{Predictor Architecture}
\label{app:architecture-details}

The predictor supports three VAE latent domains with 16, 48, and 128 channels through domain-specific $1\times1\times1$ input adapters and output heads, together with action-control adapters. The VAE-specific input maps in \cref{eq:condition-fusion} are trained jointly with the predictor to accommodate channel-count, scale, and channel-statistic differences before the shared trunk. They preserve each domain's temporal/spatial grid rather than requiring identical grids across VAEs. The future-query stream starts with a learned token and encodes preceding ground-truth slices during training or preceding predictions during forecasting. Together with the mask in \cref{eq:causal-mask}, this shift prevents access to the current target or later targets.

Each block applies Causal Temporal Attention and a feed-forward network (FFN), followed by Dense Spatial Attention and an FFN over each slice's full spatial grid. These sublayers use pre-normalization, residual connections, and positional embeddings. The trunk has four blocks, hidden width 768, 12 attention heads, an FFN expansion ratio of four, and no dropout. Domain-specific output heads map the future-query features in \cref{eq:direct-output} back to native VAE channels. The model predicts the full future latent directly; additive feedback in \cref{eq:add-feedback} modifies inputs, not outputs.

\subsection{Normalized Training Objective}
\label{app:normalized-training}

Both training stages use the normalized objective in \cref{eq:loss}. The target-dependent factors are
\begin{equation*}
  s_n=\max\!\left(\operatorname{Std}(\mathbf{z}_n),10^{-3}\right),
  \qquad
  r_n=\max\!\left(\operatorname{Std}(\Delta_T\mathbf{z}_n),10^{-3}\right),
\end{equation*}
where $s_n$ normalizes state errors, $r_n$ normalizes temporal-difference errors, and $\operatorname{Std}$ computes the population standard deviation over all entries of the corresponding target tensor. The floor prevents division by very small values. Absolute values, divisions, and squares in \cref{eq:loss} act element-wise before averaging. When $T_d=1$, no adjacent temporal slices exist and the temporal term is omitted.

Stage I adds aligned predicted and ground-truth chunks element-wise to support early optimization while retaining predicted content in the input. The sum is not averaged, and the supervision target remains the complete real future chunk. Stage II continues optimizing the same predictor after removing the direct ground-truth addition, which can change both the content and the scale of the input.
Both stages retain within-chunk teacher forcing and supervision from real future chunks. The predicted chunks used as inter-chunk feedback during training are therefore computed using ground-truth prefixes. At inference, C-NCP instead uses predicted prefixes within each chunk, and the completed video-generator output becomes the input for the next interaction. The two-stage schedule thus adapts the predictor to prediction-only inter-chunk feedback without making the training and inference distributions identical.

We optimize the predictor and its adapters with AdamW using a learning rate of $10^{-4}$, weight decay of $10^{-4}$, bfloat16 precision, gradient-norm clipping at 1.0, and a batch size of one per accelerator. The normalized absolute-error, squared-error, and temporal-difference loss weights are $\lambda_1=1$, $\lambda_2=1$, and $\lambda_{\mathrm{temp}}=0.25$, respectively. Each stage runs for 100k iterations, with the VAEs and base video generators frozen throughout.

\section{Additional Results on Sekai-val}
\label{app:sekai-results}
As shown in \cref{tab:vbench_sekai}, adding our predictor improved all six VBench dimensions and the average score for every base generator. Higher subject and background consistency accompanied improvements in temporal stability, motion smoothness, and visual quality, with LingBot-World-Fast + Ours achieving the highest average among the compared methods. These results indicate that the benefit of future-state guidance extends across different generators. A plausible explanation is that the predictor estimates how the current scene should evolve under the next action, providing a supervised visual reference that helps maintain scene content while guiding the upcoming transition.

\begin{table*}[!htbp]
\centering
\caption{Quantitative comparison on Sekai-val using VBench. All videos are 60 seconds long. Higher is better for every metric. Bold rows denote base generators augmented with our predictor. T.F.: Temporal Flickering; M.S.: Motion Smoothness; S.C.: Subject Consistency; B.C.: Background Consistency; A.Q.: Aesthetic Quality; I.Q.: Imaging Quality; Avg.: Average.}
\label{tab:vbench_sekai}
\begingroup
\resizebox{\textwidth}{!}{%
\begin{tabular}{l|cccccc|c}
\hline
\rowcolor{gray!18}
Method & T.F. $\uparrow$ & M.S. $\uparrow$ & S.C. $\uparrow$ & B.C. $\uparrow$ & A.Q. $\uparrow$ & I.Q. $\uparrow$ & Avg. $\uparrow$ \\
\hline
YUME 1.5 & 0.9002 & 0.9321 & 0.7115 & 0.8921 & 0.4270 & 0.6877 & 0.7584 \\
FantasyWorld & 0.9024 & 0.9298 & 0.7162 & 0.8905 & 0.4291 & 0.6849 & 0.7588 \\
Infinite-World & 0.8987 & 0.9340 & 0.7089 & 0.8952 & 0.4258 & 0.6905 & 0.7588 \\
\hline
SANA-WM & 0.9413 & 0.9713 & 0.7364 & 0.9119 & 0.4522 & 0.7237 & 0.7895 \\
\textbf{SANA-WM + Ours} & \textbf{0.9534} & \textbf{0.9850} & \textbf{0.7804} & \textbf{0.9288} & \textbf{0.4743} & \textbf{0.7371} & \textbf{0.8098} \\
\hline
Matrix-game-3.0 & 0.9360 & 0.9708 & 0.7600 & 0.9104 & 0.4535 & 0.7555 & 0.7977 \\
\textbf{Matrix-game-3.0 + Ours} & \textbf{0.9482} & \textbf{0.9843} & \textbf{0.8022} & \textbf{0.9273} & \textbf{0.4663} & \textbf{0.7689} & \textbf{0.8162} \\
\hline
HY-World 1.5 & 0.9319 & 0.9718 & 0.7209 & 0.9148 & 0.4303 & 0.7014 & 0.7785 \\
\textbf{HY-World 1.5 + Ours} & \textbf{0.9440} & \textbf{0.9853} & \textbf{0.7609} & \textbf{0.9318} & \textbf{0.4424} & \textbf{0.7138} & \textbf{0.7964} \\
\hline
LingBot-World-Fast & 0.9560 & 0.9713 & 0.7284 & 0.9104 & 0.4729 & 0.7142 & 0.7922 \\
\textbf{LingBot-World-Fast + Ours} & \textbf{0.9679} & \textbf{0.9854} & \textbf{0.7781} & \textbf{0.9273} & \textbf{0.5141} & \textbf{0.7286} & \textbf{0.8169} \\
\hline
\end{tabular}%
}
\endgroup
\end{table*}

\section{Additional Results on WBench}
\label{app:wbench-results}
On the official WBench test set, our predictor improved Quality, Setting, Interaction, Consistency, and Physical scores for all four generators, as shown in \cref{tab:wbench}. Consistency showed the largest gain within each pair, while HY-World 1.5 + Ours achieved the highest average among the compared methods. The joint improvements in interaction and consistency suggest that future-state guidance helps preserve scene content while responding to action controls. This trend is consistent with repeatedly predicting the next state from the current generated latent and the upcoming action, then using the forecast to refine generation before its output becomes the context for another interaction.

\begin{table}[h]
\centering
\caption{Quantitative comparison on the official WBench test set using 60-second videos. Higher is better for every metric. Bold rows denote base generators augmented with our predictor. Qual.: Quality; Set.: Setting; Inter.: Interaction; Cons.: Consistency; Phys.: Physical; Avg.: Average.}
\label{tab:wbench}
\begingroup
\resizebox{\textwidth}{!}{%
\begin{tabular}{l|ccccc|c}
\hline
\rowcolor{gray!18}
Method & Qual. $\uparrow$ & Set. $\uparrow$ & Inter. $\uparrow$ & Cons. $\uparrow$ & Phys. $\uparrow$ & Avg. $\uparrow$ \\
\hline
YUME 1.5 & 77.6 & 72.4 & 72.0 & 80.1 & 65.2 & 73.5 \\
FantasyWorld & 72.4 & 71.3 & 72.1 & 86.4 & 66.8 & 73.8 \\
Infinite-World & 77.0 & 69.3 & 75.9 & 80.0 & 62.1 & 72.9 \\
DreamX-World & 77.5 & 80.8 & 78.4 & 74.9 & 63.3 & 75.0 \\
\hline
SANA-WM & 79.3 & 76.1 & 82.1 & 80.7 & 61.9 & 76.0 \\
\textbf{SANA-WM + Ours} & \textbf{80.3} & \textbf{77.1} & \textbf{83.7} & \textbf{85.2} & \textbf{63.1} & \textbf{77.9} \\
\hline
Matrix-game-3.0 & 75.5 & 63.6 & 83.5 & 74.5 & 59.3 & 71.3 \\
\textbf{Matrix-game-3.0 + Ours} & \textbf{76.5} & \textbf{64.5} & \textbf{85.1} & \textbf{78.6} & \textbf{60.4} & \textbf{73.0} \\
\hline
HY-World 1.5 & 78.1 & 72.2 & 87.5 & 86.9 & 66.3 & 78.2 \\
\textbf{HY-World 1.5 + Ours} & \textbf{79.1} & \textbf{73.2} & \textbf{89.1} & \textbf{91.7} & \textbf{67.5} & \textbf{80.1} \\
\hline
LingBot-World-Fast & 79.4 & 77.9 & 79.4 & 84.9 & 65.7 & 77.5 \\
\textbf{LingBot-World-Fast + Ours} & \textbf{80.4} & \textbf{79.0} & \textbf{80.9} & \textbf{89.6} & \textbf{66.9} & \textbf{79.4} \\
\hline
\end{tabular}%
}
\endgroup
\end{table}

\section{Training Strategy Ablation}
\label{app:training-strategy-ablation}

\begin{table}[h]
\centering
\caption{Training-strategy ablation with Matrix-game-3.0 on RealEstate10K-test using VBench. All videos are 60 seconds long. Stage inputs specify inter-chunk feedback; Pred+GT denotes element-wise addition. Higher is better for every metric, and the full method is highlighted in bold. T.F.: Temporal Flickering; M.S.: Motion Smoothness; S.C.: Subject Consistency; B.C.: Background Consistency; A.Q.: Aesthetic Quality; I.Q.: Imaging Quality; Avg.: Average.}
\label{tab:training_strategy_ablation}
\resizebox{\linewidth}{!}{%
\begin{tabular}{l|cc|cccccc|c}
\hline
\rowcolor{gray!18}
Variant & Stage I Input & Stage II Input & T.F. $\uparrow$ & M.S. $\uparrow$ & S.C. $\uparrow$ & B.C. $\uparrow$ & A.Q. $\uparrow$ & I.Q. $\uparrow$ & Avg. $\uparrow$ \\
\hline
One-stage Pred & Pred & -- & 0.9290 & 0.9785 & 0.8110 & 0.9110 & 0.5050 & 0.7831 & 0.8196 \\
One-stage GT & GT & -- & 0.9315 & 0.9825 & 0.8205 & 0.9145 & 0.5090 & 0.7840 & 0.8237 \\
One-stage Pred+GT & Pred+GT & -- & 0.9330 & 0.9842 & 0.8268 & 0.9168 & 0.5118 & 0.7846 & 0.8262 \\
Two-stage GT$\rightarrow$Pred & GT & Pred & 0.9345 & 0.9860 & 0.8320 & 0.9185 & 0.5135 & 0.7860 & 0.8284 \\
\hline
\textbf{Full Ours} & Pred+GT & Pred & \textbf{0.9361} & \textbf{0.9884} & \textbf{0.8387} & \textbf{0.9210} & \textbf{0.5167} & \textbf{0.7877} & \textbf{0.8314} \\
\hline
\end{tabular}%
}
\end{table}

\Cref{tab:training_strategy_ablation} compares five inter-chunk input schedules; within-chunk teacher forcing is retained. One-stage Pred achieved the lowest scores, consistent with errors in early predictions making subsequent inputs less reliable. One-stage GT performed better, but dependence on real latent inputs leaves a mismatch with prediction-based feedback. One-stage Pred+GT improved further by exposing the predictor to its own outputs alongside ground-truth information, although it never adapts to prediction-only inputs. Two-stage GT$\rightarrow$Pred outperformed all one-stage variants, supporting a transition from ground-truth-supported learning to prediction-dependent training. Full Ours achieved the highest score on every metric. Its first stage combines predicted and real chunks, providing ground-truth support while exposing the model to its own prediction errors; the second removes inter-chunk ground-truth addition to adapt the predictor to its own feedback. These results support combining a ground-truth-supported warm-up with prediction-only adaptation for more reliable next-state prediction.

\section{Additional Qualitative Results}
\label{app:additional-qualitative}

\begin{figure}[h]
  \centering
  \includegraphics[width=\textwidth]{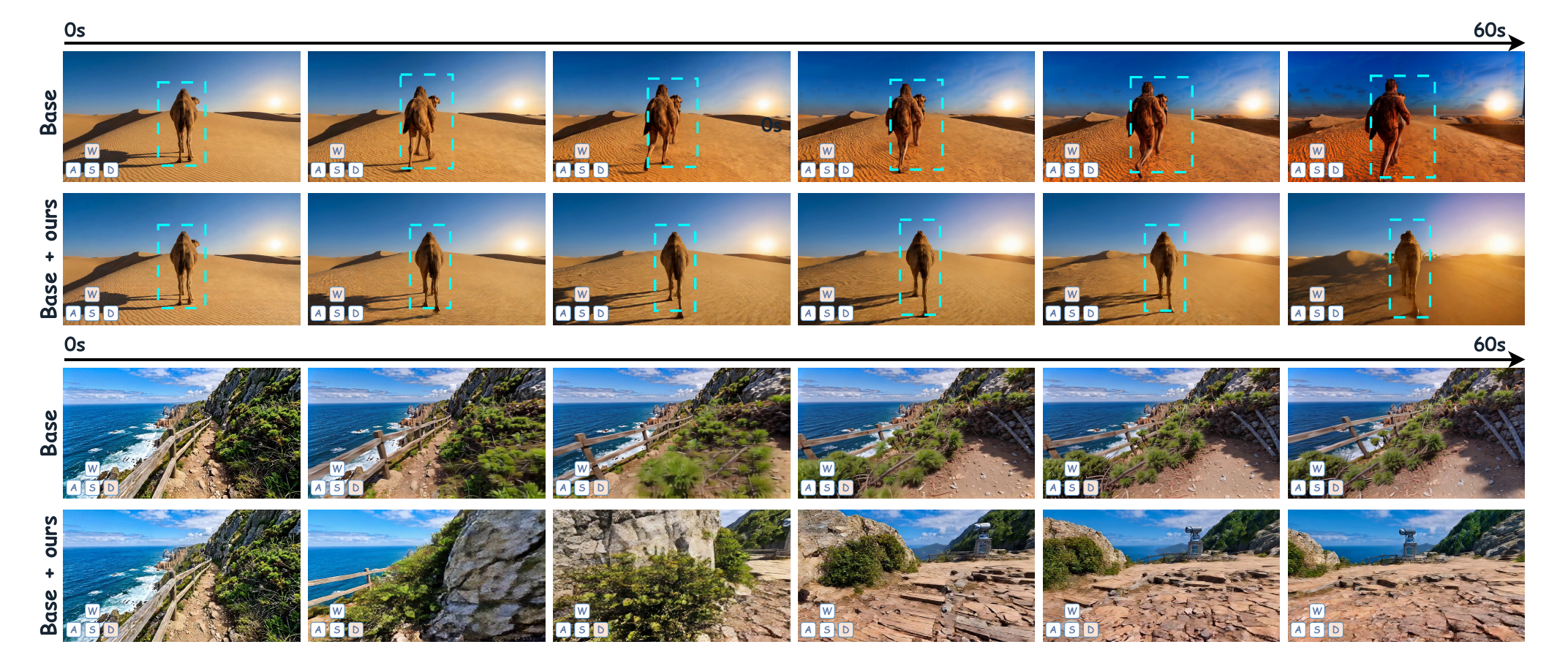}
  \caption{Additional qualitative comparisons over 60-second interactive rollouts in desert and coastal scenes. Each pair shows the base generator above and its predictor-augmented counterpart below, with six snapshots ordered from left to right. Action controls appear in the lower-left corners, and cyan dashed boxes highlight the camel in the desert example.}
  \label{fig:qualitative-additional}
\end{figure}

\Cref{fig:qualitative-additional} provides two additional examples of sustained interactive generation. In the desert scene, the baseline gradually changed the camel's body shape and appearance and introduced irregular sky textures, whereas the predictor-augmented sequence retained a more stable camel silhouette and scene appearance. In the coastal scene, later baseline frames remained around a similar section of the trail, with less distinct foreground vegetation and path details. With our predictor, the sequence progressed along the rocky path toward an open viewpoint while preserving recognizable rock surfaces and path boundaries. These examples illustrate the value of future-state guidance for maintaining subject appearance while allowing the scene to evolve during interaction. By estimating the upcoming state from the current generated latent and the next action, the predictor provides a future-state prior for each new chunk. Modulation uses this prior to refine generation, helping limit the continuation of visual defects without simply preserving the current view.

\begin{figure}[t]
  \centering
  \includegraphics[width=\textwidth]{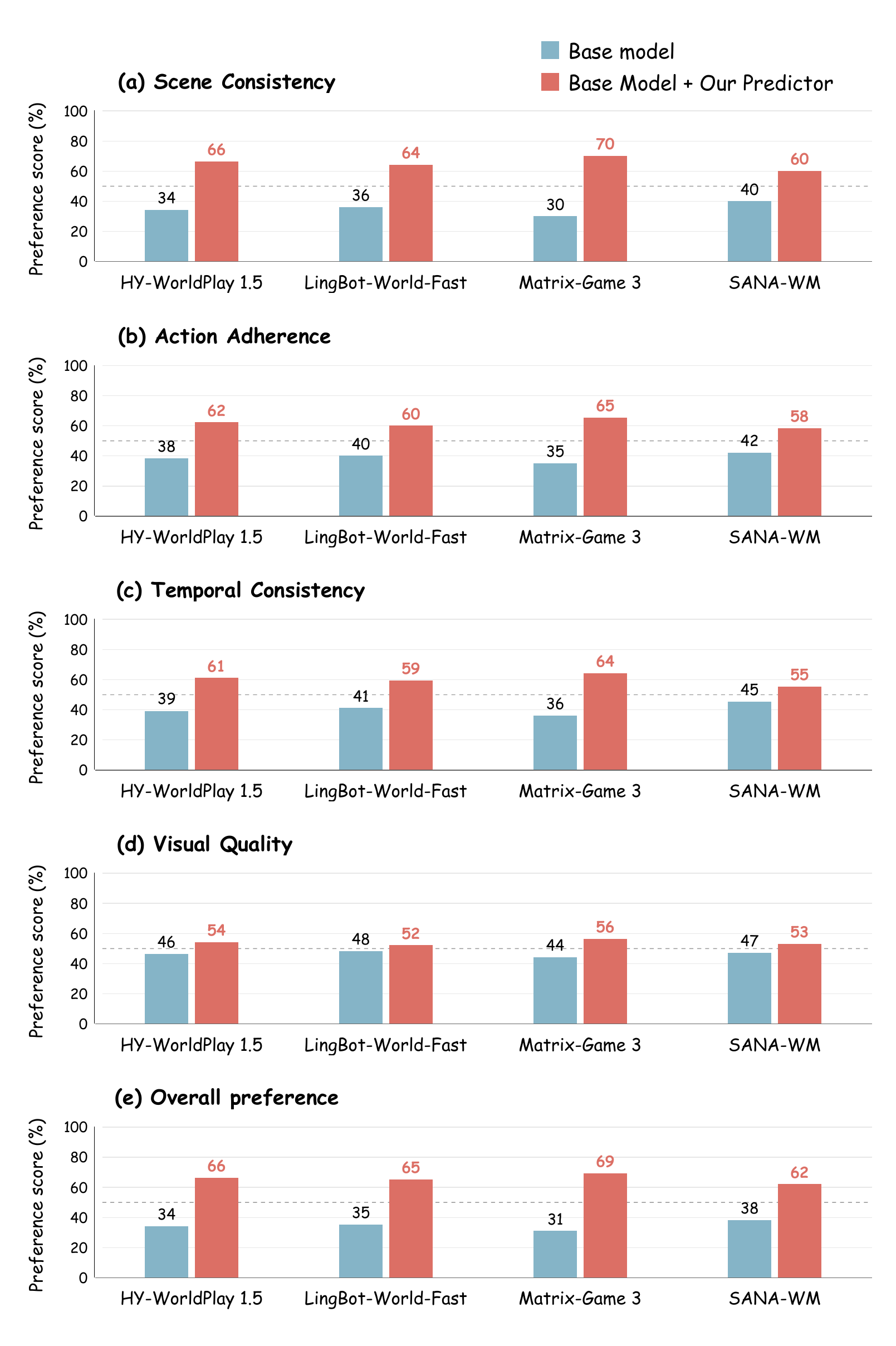}
  \caption{User-study preferences across four base generators on scene consistency, action adherence, temporal consistency, visual quality, and overall preference. Blue and red bars show the preference percentages for the base model and its predictor-augmented counterpart, respectively. The dashed line marks equal preference at 50\%.}
  \label{fig:user-study}
\end{figure}

\section{User Study}
\label{app:user-study}
We conducted a paired user study with 100 participants, each evaluating all 20 video pairs, with five pairs per base generator. 
The study covered five criteria. \emph{Scene consistency} assessed whether object appearance, position, and scene layout remained consistent through turns, return paths, and revisits. \emph{Action adherence} assessed whether the generated motion followed the displayed controls, including forward and backward movement, lateral movement, and turning. \emph{Temporal consistency} assessed flickering, abrupt changes, and continuity across action switches and chunk boundaries. \emph{Visual quality} assessed clarity, texture, structural integrity, and visible distortions or artifacts. \emph{Overall preference} captured which result participants preferred after considering these aspects together.

As shown in \cref{fig:user-study}, participants preferred predictor-augmented videos on all five criteria for all four generators. This preference was stronger for scene consistency, action adherence, and temporal consistency, while visual quality showed a more modest advantage. The pattern is consistent with the predictor's role in anticipating scene transitions. By estimating the next state from the current generated latent and the next action, the predictor supplies a visual reference for how the scene should evolve. Modulation uses this reference to refine each new chunk before it becomes part of the history. The preference pattern thus supports future-state guidance for preserving coherent, action-responsive scenes during sustained interaction.

\end{document}